\documentclass{article}

\usepackage{arxiv}

\usepackage[utf8]{inputenc} 
\usepackage[T1]{fontenc}    
\usepackage{hyperref}       
\usepackage{url}            
\usepackage{booktabs}       
\usepackage{amsfonts}       
\usepackage{nicefrac}       
\usepackage{microtype}      
\usepackage{lipsum}
\usepackage{graphicx}
\usepackage{indentfirst}
\graphicspath{ {./images/} }
\usepackage[table,xcdraw]{xcolor}
\usepackage{enumerate}
\usepackage[shortlabels]{enumitem}

\usepackage{authblk}

\usepackage{wrapfig}
\usepackage{graphicx}
\usepackage{picinpar}

\makeatletter
\long\def\figwindownonum[#1,#2,#3,#4] {
  \begin{window}[#1,#2,{#3},{\centering#4\par}] }
\def\endfigwindownonum{\end{window}}
\makeatother

\title{The 2nd Anti-UAV Workshop \& Challenge: \\
Methods and Results}

\author[1]{Jian Zhao}
\author[2]{Gang Wang}
\author[3]{Jianan Li}
\author[4]{Lei Jin}
\author[5]{Nana Fan}
\author[6]{Min Wang}
\author[4]{Xiaojuan Wang}
\author[1]{Ting Yong}
\author[7]{Yafeng Deng}
\author[8]{Yandong Guo}
\author[9]{Shiming Ge}
\author[10]{Guodong Guo}
\affil[1]{Institute of North Electronic Equipment}
\affil[2]{Beijing Institute of Basic Medical Sciences}
\affil[3]{Beijing Institute of Technology}
\affil[4]{Beijing University of Posts and Telecommunication}
\affil[5]{Harbin Institute of Technology}
\affil[6]{Beijing Jiaotong University}
\affil[7]{Qihoo 360}
\affil[8]{OPPO Research Institute}
\affil[9]{Institute of Information Engineering, CAS}
\affil[10]{Institute of Deep Learning, Baidu Research}

\begin{document}
\maketitle
\begin{abstract}
The 2nd Anti-UAV Workshop \& Challenge aims to encourage research in developing novel and accurate methods for multi-scale object tracking. The Anti-UAV dataset used for the Anti-UAV Challenge has been publicly released. There are two subsets in the dataset, $i.e.$, the test-dev subset and test-challenge subset. Both subsets consist of 140 thermal infrared video sequences, spanning multiple occurrences of multi-scale UAVs. Around 24 participating teams from the globe competed in the 2nd Anti-UAV Challenge. In this paper, we provide a brief summary of the
2nd Anti-UAV Workshop \& Challenge including brief introductions to the top three methods.The submission leaderboard will be reopened for researchers that are
interested in the Anti-UAV challenge. The benchmark dataset and other information can be found at: https://anti-uav.github.io/.
\end{abstract}


\section{Introduction}

Civil unmanned aerial vehicle (UAV) is growing rapidly in a wide range of consumer communications and networks with their autonomy, flexibility, and a broad range of application domains. UAV applications offer possible civil and public domain applications in which single or multiple UAVs may be used. Nevertheless, we should also be aware of the potential threat to our lives caused by UAV intrusion, since UAVs can also be used to conduct physical attacks ($e.g.$, via explosives) and cyber-attacks ($e.g.$, hacking a critical infrastructure). Moreover, unauthorized UAVs are a danger to civilian aircraft. There have been multiple instances of drone sightings that halted air traffic at airports, leading to significant economic losses for airlines.

Historically, radar is certainly a very powerful technology for detecting traditional incoming airborne threats. However, these comparatively small drones are difficult for radar to accurately detect, because they have very small radar cross-sections and erratic flight paths. Therefore, how to use computer vision algorithms to perceive UAVs is a crucial part of the whole UAV-defense system.

\begin{figure}[h] 
\vspace{-2mm}
\begin{center}
\includegraphics[width=0.85\linewidth]{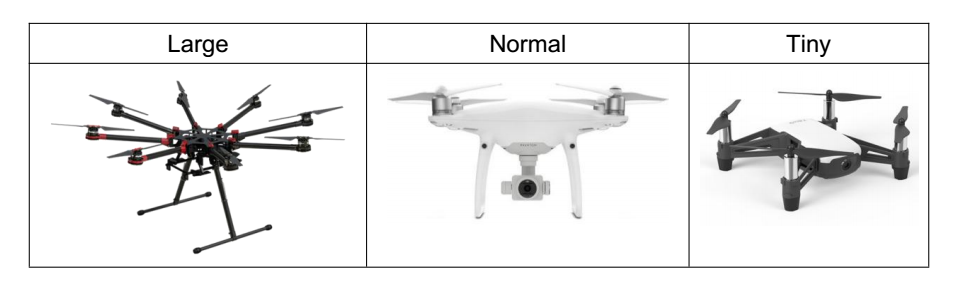}
\end{center}
   \caption{Illustrations of civil UAVs: Large civil UAV; Small civil UAV; Tiny civil UAV.}
\label{fig:1}
\end{figure}

Traditional computer vision research~\cite{Marginalized CNN, Dynamic Conditional Networks,re10,re11,re12} for UAV detection and tracking lacks a high-quality benchmark in dynamic environments. To mitigate this gap, we held the 1st International Workshop on Anti-UAV Challenge~\cite{re13} at CVPR 2020, releasing a dataset consisting of 160 video sequences (both RGB and infrared). The workshop attracted attention from researchers all over the world. Many submitted solutions outperform the baseline method, making great contributions to addressing the anti-UAV problem~\cite{re13,re14,re15}. The 2nd anti-UAV challenge extends the benchmark dataset to 280 high-quality, full HD thermal infrared video sequences, spanning multiple occurrences of multi-scale ($i.e.$, large, small and tiny, as shown in Fig.\ref{fig:1}) UAVs. The workshop encourages participants to develop automated methods that can detect and track UAVs in thermal infrared videos with high accuracy. Particularly, algorithms that can detect and track fast-moving drones in complex environments ($e.g.$, occlusion by cloud/buildings/trees, and fake targets like kites, balloons, birds, etc.) are highly expected.

This workshop will bring together academic and industrial experts in the field of UAVs to discuss the techniques and applications of tracking UAVs. Participants are invited to submit their original contributions, surveys, and case studies that address the works of UAV’s detection and tracking issues.

\section{The ANTI-UAV Challenge}
\subsection{Dataset}
There are two subsets in the dataset, $i.e.$, the test-dev subset and test-challenge subset. Both subsets consist of 140 thermal infrared video sequences, spanning multiple occurrences of multi-scale UAVs. We only provide annotation files for the test-dev. Compared to the previous challenge, we enlarge both the test-dev and the test-challenge in this year by adding more challenging video sequences with dynamic backgrounds and small-scale targets, such that the resulting new dataset covers a greater variety of scenarios with multi-scale UAVs. The target scales include large, medium, small, and tiny. Besides, The videos recorded include two lighting conditions (day and night), and diverse backgrounds (buildings, cloud, trees, sea, $etc$.). 

\subsection{Metric}
Anti-UAV is annotated with bounding boxes, attributes and
existing flags. Moreover, an empty bounding box list denotes a ”not exist” flag. Trackers need to obtain the perception of UAV status. In this case, the presence of UAV in the visual range is introduced into the evaluation metric:
\begin{equation}
    acc=\sum_{t=1}^T{\frac{IoU_t\times 1\left[ v_t>0 \right] +p_t\times \left( 1-1\left[ v_t>0 \right] \right)}{T}},
\end{equation}
For frame $t$, $IoU_t$ is Intersection over Union (IoU) between the predicted tracking box and its corresponding ground-truth box, $p_t$ equals 1 when the predicted box is empty and 0 otherwise, and $v_t$ is the ground-truth existence/visibility flag of the target. The Iverson bracket indicator function $1[v_t>0]$ equals 1 when $[v_t>0]$ and 0 otherwise. The accuracy is averaged over all $T$ frames.

\section{Result and Method}
The 2nd Anti-UAV challenge was held between May 12, 2021 and July 10, 2021.The results of the 2nd Anti-UAV challenge are shown in Table 1. Around 24 teams submitted their final results in this challenge. In this section,we will briefly introduce the methodologies of the top 3 submissions.

\begin{table}[]
\centering
\caption{Challenge results}
\label{tab1}
\begin{tabular}{
>{\columncolor[HTML]{FFFFFF}}c 
>{\columncolor[HTML]{FFFFFF}}c 
>{\columncolor[HTML]{FFFFFF}}c }
\hline
{\color[HTML]{333333} \textbf{Rank}} & {\color[HTML]{333333} \textbf{User Name}}             & {\color[HTML]{333333} \textbf{Tracking Accuracy}} \\ \hline
{\color[HTML]{333333} 1}             & {\color[HTML]{333333} huangbo940326}             & {\color[HTML]{333333} 0.6444}                     \\
{\color[HTML]{333333} 2}             & {\color[HTML]{333333} whoamiw}                   & {\color[HTML]{333333} 0.6388}                     \\
{\color[HTML]{333333} 3}             & {\color[HTML]{333333} JUN}                       & {\color[HTML]{333333} 0.6380}                     \\
{\color[HTML]{333333} 4}             & {\color[HTML]{333333} tang\_god}                 & {\color[HTML]{333333} 0.6356}                     \\
{\color[HTML]{333333} 5}             & {\color[HTML]{333333} ZhangyongTang}             & {\color[HTML]{333333} 0.6215}                     \\
{\color[HTML]{333333} 6}             & {\color[HTML]{333333} QLY}                       & {\color[HTML]{333333} 0.6130}                     \\
{\color[HTML]{333333} 7}             & {\color[HTML]{333333} jjchen}                    & {\color[HTML]{333333} 0.6116}                     \\
{\color[HTML]{333333} 8}             & {\color[HTML]{333333} blue\_star}                & {\color[HTML]{333333} 0.6116}                     \\
{\color[HTML]{333333} 9}             & {\color[HTML]{333333} hli1221}                   & {\color[HTML]{333333} 0.6082}                     \\
{\color[HTML]{333333} 10}            & {\color[HTML]{333333} zhangxiaohan\_zhaojinjian} & {\color[HTML]{333333} 0.6066}                     \\
{\color[HTML]{333333} 11}            & {\color[HTML]{333333} YouKnowWhoAmI}             & {\color[HTML]{333333} 0.6062}                     \\
{\color[HTML]{333333} 12}            & {\color[HTML]{333333} leili}                     & {\color[HTML]{333333} 0.5848}                     \\
{\color[HTML]{333333} 13}            & {\color[HTML]{333333} guyu}                      & {\color[HTML]{333333} 0.5824}                     \\
{\color[HTML]{333333} 14}            & {\color[HTML]{333333} jinke}                     & {\color[HTML]{333333} 0.5795}                     \\
{\color[HTML]{333333} 15}            & {\color[HTML]{333333} xjtudz}                    & {\color[HTML]{333333} 0.5752}                     \\
{\color[HTML]{333333} 16}            & {\color[HTML]{333333} shan666}                   & {\color[HTML]{333333} 0.5681}                     \\
{\color[HTML]{333333} 17}            & {\color[HTML]{333333} adamzdw}                   & {\color[HTML]{333333} 0.5603}                     \\
{\color[HTML]{333333} 18}            & {\color[HTML]{333333} Homura}                    & {\color[HTML]{333333} 0.5544}                     \\
{\color[HTML]{333333} 19}            & {\color[HTML]{333333} jkahsjkd}                  & {\color[HTML]{333333} 0.5251}                     \\
{\color[HTML]{333333} 20}            & {\color[HTML]{333333} ywang26}                   & {\color[HTML]{333333} 0.5163}                     \\
{\color[HTML]{333333} 21}            & {\color[HTML]{333333} Nitre}                     & {\color[HTML]{333333} 0.5139}                     \\
{\color[HTML]{333333} 22}            & {\color[HTML]{333333} zhuwenming}                & {\color[HTML]{333333} 0.5057}                     \\
{\color[HTML]{333333} 23}            & {\color[HTML]{333333} tangyuan23}                & {\color[HTML]{333333} 0.4941}                     \\
{\color[HTML]{333333} 24}            & {\color[HTML]{333333} kostadinov}                & {\color[HTML]{333333} 0.4930}                     \\ \hline
\end{tabular}
\end{table}

\subsection{Team BIT\_OITS}
\textbf{Bo Huang, Junjie Chen, Shenwang Jiang, Ying Wang, Yuncheng Wang, Lei Wang, Tingfa Xu.} (Beijing Institute of Technology (BIT) \& Beijing Institute of Technology Chongqing Innovation Center (BITCQIC))

The authors propose a robust spatio-temporal attention based Siamese (SiamSTA) tracker to track UAV targets in thermal infrared (TIR) videos. The SiamSTA tracker is built on the Siam R-CNN~\cite{SiamR-CNN} network, and they borrow the pre-trained weights from Siam R-CNN. Like other Siamese frameworks, SiamSTA also consists of two branches: a template branch that is initialized by the first frame and a test branch that feeds the current detecting image. These two branches share the same weights, and are followed by a two-stage re-detector to compute the confidence scores for the enumeration of multiple RPN proposals.

\textbf{Siam R-CNN.} In Siam R-CNN, the authors develop a tracklet dynamic programming algorithm (TDPA) to implicitly track both the object of interest and potential similar-looking distractors using spatio-temporal cues. A tracklet means a trajectory which consists of a sequence of non-overlapping rajectories. Siam R-CNN backs up a lot of such tracklets to filter the optimal predicted bounding box from thousands of candidate proposals. The same knife cuts bread and fingers.

\textbf{SiamSTA Tracker Design.} For a tracking task in TIR videos, the object is often textureless, especially for far-range drone planar argets. In some extreme cases, the drone target is very small, even like a point target, and there will be a plethora of distractors in the scene. At this time, these tracklets may bring additional interference due to the difficulty of extracting high-quality semantic features for the weak UAV targets. To address such issue, finer exploitation of spatio-temporal attention mechanisms is a feasible solution. The SiamSTA tracker is designed as following: Firstly, the tracker records the target's size, ratio for all previous frames, and the new predicted tracklets must be within the range of [0.8 * min(size), 1.2 * max(size)], [0.8 * min(ratio), 1.2 * max(ratio)]. Other tracklets are treated as background distractors and will be terminated directly. The authors then develop an optical flow-based algorithm to estimate global background motion. They use the ShiTomasi algorithm to compute the key points, and constrain the number of key points in the range of 5-30. They then apply the Lucas-Kanade (L-K) optical flow to track these key points, and select the points whose forward-backward (F-B) error is less than 1. If the average location diff of these selected key points is less than 0.5 for 5 consecutive frames, they consider the camera to be static. Under a long-range static camera, they assume that there are no huge position jumps for these targets between two adjacent frames, thus they can obtain a more reliable tracking result by searching the nearby area. In this case, authors set the overlap of two consecutive tracklets to be greater than 0.1, they will delete the other tracklets.

\textbf{Model Refinement.} However, if the target is lost, using the nearby detecting strategy will cause the model to fail completely. In order to cope with the target loss, especially when handling the challenge of target occlusion or out of view, a mature global re-detection becomes necessary to recover the tracking failures. Therefore, it is very critical to define the boundaries of nearby tracking and global re-detection to improve tracking performance. If the background is dynamic and the re-detection score is greater than 0, authors output the tracklet with the best score. On the contrast, they implement the change detection algorithms from the pybgs library: one is FrameDifference, the other is DPGrimsonGMM. If the output of tracklets, FrameDifference and DPGrimsonGMM have a large overlap, they believe that the tracking result is surely correct, and they will use nearby detecting in the next frame. If the output of FrameDifference and DPGrimsonGMM have a large overlap, but the location of tracklets is isolated, authors believe that the tracklets fail and add a new tracklet initialized by moving target detection. If there is no overlap for these three tracking results, authors output the bounding box of tracklet with the best score, and they will perform a global re-detection in the next frame.

\textbf{Ablation study.} Authors do ablation experiments to verify the effect of each incremental component, including lost definition, change detection (CD), and spatio temporal attention (STA). They provide six sets of tracking results which are described as follows: 
\begin{enumerate}[-,nosep]
    \item Siam RCNN Baseline: it is the same method as Siam RCNN without changing anything.
    \item Siam RCNN Lost: they add a lost definition to Siam RCNN, if the score of the best tracklet is less than 0.0, they consider the tracking to be a failure and output the empty result.
    \item Siam RCNN + STA: they add the spatio temporal constraints mentioned above to Siam RCNN. In addition, if the score of the best tracklet is greater than 0.9, they perform a nearby searching, and otherwise they will do a global redetection. 
    \item Siam RCNN + CD: they utilize the change detection method to re capture the target, tracklets with a large overlap with change detection results will be assigned a high score, and untrustworthy tracklets will be deleted directly.
    \item CD + STA: this is a pure motion detection tracker without Siam RCNN. To cope with the background motion and target jump, they add the same spatio temporal constraints and introduce a correlation filter based algorithm to assist the tracking.
    \item Siam RCNN + STA + CD: this is their final method as previously stated.
\end{enumerate}

\textbf{Main contribution.} Authors propose several practical guidelines to solve the challenge of drone tracking, such as weak targets, small targets.
Their tracker makes full use of prior knowledge like scale/ratio distribution by combining neighborhood search and global detection. The proposed CD+STA algorithm fully exploits valuable motion information, which can greatly benefit the tacking of tiny moving drone targets.

\subsection{Team COLA Try}
\textbf{Zitian Wang, Shangzhe Di, Zongheng Tang, Si Liu.} (Beihang University)

Authors adapted some recent trackers (like ECO~\cite{ECO}, KYS, SuperDiMP~\cite{bhat2019learning} and Stark) to the LTMU~\cite{LTMU} long-term tracking framework, with pluggable AlphaRef~\cite{AlphaRef} for scale estimation. These trackers are built for producing more robust tracking results through the multi-tracker voting and fusion. Specifically, for each frame, all the trackers first produce the tracking results respectively, then an IOU-based clustering algorithm is applied to generate a set of clusters. The box and confidence of each cluster are reweighted by the elements belonging to this cluster and the cluster size. The cluster with the highest confidence will serve as the final prediction of this frame. They also adopted the motion enhancement to distinguish small UAVs from background noises. First, optical flow is predicted between adjacent frames. Optical flow mainly captures the camera movement between adjacent frames, while ignoring the UAV movement due to the very small size. Then the image warping based on optical flow is to make the background in adjacent frames more aligned. So they use a background subtraction method to capture the motion pattern of the two frames. The output motion activation map serve as prior knowledge which indicates the moving small UAVs, and is fused with original frame for motion enhancement.

\textbf{Ablation study.} 
\begin{enumerate}[-,nosep]
    \item SuperDiMP + AlphaRef  -> 60.6 
    \item SuperDiMP + AlphaRef + LTMU -> 64.0
    \item Stark -> 60.6 
    \item Stark + AlphaRef -> 62.1
    \item Stark + AlphaRef + LTMU -> 63.1
    \item multi-tracker voting and fusion -> 67.5
    \item +motion enhancement -> 68.7
\end{enumerate}

\textbf{Main contribution.}
Since the organizers require that "test-dev" data should not be used for training and no detector should be applied, authors consider the problem from the perspective of exploiting trackers trained in another domain (RGB input, general scenes) to track UAVs with IR input. They proposed multi-tracker voting and fusion (can be seen as one kind of ensemble) to make the most of different tracking mechanisms and generate more robust tracking results.  On the other hand, the pretrained trackers struggled to track the small UAVs with background noises due to unseen scenarios and poor appearance information. The motion enhancement method is meant to better find these small UAVs using the motion information. 

\subsection{Team JNU}
\textbf{Xuefeng Zhu, Zhangyong Tang, Hui Li, Tianyang Xu, Xiaojun Wu, Josef Kittler.} (Jiangnan University \& University of Surrey)

This tracker is improved based on SuperDiMP~\cite{bhat2019learning}, SiamRPN++~\cite{li2019siamrpn++} and TransT~\cite{chen2021transformer} methods. Above all, the target state is predicted normally by detecting multi-scale search regions using a local tracker that combines SuperDiMP, Transt, and SiamRPN++ methods. Then the local result is verified by taking into account the predicted target appearance and response scores of the local tracker. If the verification result indicates that the result is correct, the normal local tracking is conducted in next frame. Otherwise, the re-detection module will be activated. Authors adopt a violent method to search the whole image by sliding windows, to determine whether the target is present and to recapture the lost target by detecting each sliding window if the target exists. Besides, the motion information provided by optical flow is also employed for target re-detection.

\textbf{Ablation study.} 
\begin{enumerate}[-,nosep]
    \item superDiMP  -> 56.3 
    \item Multi-scale superDiMP -> 61.1
    \item Multi-scale superDiMP + SiamPRN++ -> 62.7 
    \item Multi-scale superDiMP + TansT + SiamRPN++ -> 64.0
    \item Multi-scale SuperDiMP + TansT+ SiamRPN++ + Re-detection -> 67.9
\end{enumerate}

\section{Conclusions}
Object detection and tracking in the wild scenarios are fundamental yet challenging problems in computer vision. We held the 2nd Anti-UAV Challenge to encourage researchers from the fields of object detection, visual tracking and other disciplines to present their progress, communication and novel ideas that will potentially shape the future of the UAV detection area. Approximately 24 teams around the globe participated in this competition, in which top-3 leading teams, together with their methods, are briefly introduced in this paper. Our workshop takes a different perspective, making UAVs as tracking targets, and provides a large-scale dataset to promote deep network learning for UAVs. In addition, the proposed workshop also aims at tiny object detection and tracking in the wild which is more challenging, more practical, and more useful for real applications. Thus, our workshop will bridge the needs of industry and research in academia, and may accelerate the process of these computer vision technologies being used in real applications.

\section{Acknowledgement}
This work was partially supported by the National Science Foundation of China 62006244.

\bibliographystyle{unsrt}  


\vspace*{3\baselineskip}

\begin{figwindownonum}[0,l,{\mbox{\includegraphics[width=3cm]{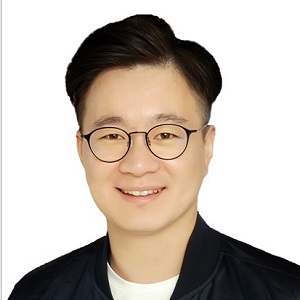}}},{}]
\noindent
\textbf{Jian Zhao}  is currently an Assistant Professor with Institute of North Electronic Equipment, Beijing, China. He received his Ph.D. degree from National University of Singapore (NUS) in 2019 under the supervision of Assist. Prof. Jiashi Feng and Assoc. Prof. Shuicheng Yan. He is the SAC of VALSE, the committee member of CSIG-BVD, and the member of the board of directors of BSIG. He has received the "2020-2022 Youth Talent Promotion Project" from China Association for Science and Technology, and the "2021-2023 Beijing Youth Talent Promotion Project" from Beijing Association for Science and Technology. He has published over 40 cutting-edge papers on human-centric image understanding. He has won the Lee Hwee Kuan Award (Gold Award) on PREMIA 2019 and the "Best Student Paper Award" on ACM MM 2018 as the first author. He has received the nomination for the USERN Prize 2021, according to publications as first author in top rank (Q1) journals of the field of Artificial Intelligence, Pattern Recognition, Machine Learning, Computer Vision and Multimedia Analytics, in the recent two years. He has won the top-3 awards several times on world-wide competitions on face recognition, human parsing and pose estimation as the first author. His main research interests include deep learning, pattern recognition, computer vision and multimedia. He and his collaborators has also successfully organized the CVPR 2020 Anti-UAV Workshop and Challenge, the ECCV 2020 RLQ-TOD Workshop and Challenge, and the CVPR 2018 L.I.P Workshop and MHP Challenge.
\end{figwindownonum}

\vspace{0.5 cm}

\begin{figwindownonum}[0,l,{\mbox{\includegraphics[width=3cm]{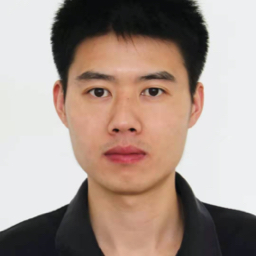}}},{}]
\noindent
\textbf{Gang Wang} is currently an Assistant Professor with Center of Brain Sciences, Beijing Institute of Basic Medical Sciences, Beijing, China. He received the Ph.D. degree from Ghent University, Belgium in 2019. He is the first author of more than 15 papers published in conference proceedings (e.g., ICCV) and international journals (e.g., IEEE-TIP). He won the best student paper nomination prizes at the EUSFLAT2017 and BNAIC2019. He is also a member of IEEE and CSCS. His research interests include computer vision, machine learning and brain-inspired vision computing. 
\end{figwindownonum}

\vspace{0.1 cm}

\begin{figwindownonum}[0,l,{\mbox{\includegraphics[width=3cm]{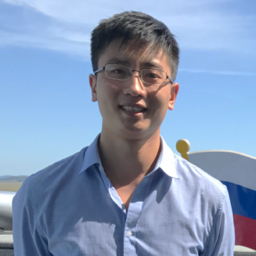}}},{}]
\noindent
\textbf{Jianan Li} is currently an Assistant Professor at School of Optoelectronics, Beijing Institute of Technology, Beijing, China, where he received his B.S. and Ph.D. degree in 2013 and 2019, respectively. From July 2019 to July 2020, he worked as a research fellow at National University of Singapore, where he also worked as a joint training Ph.D. student from July 2015 to July 2017. From October 2017 to April 2018, he worked as an intern at Adobe Research. His research interests mainly include computer vision and real-time image/video processing.
\end{figwindownonum}
\vspace{0.5 cm}

\begin{figwindownonum}[0,l,{\mbox{\includegraphics[width=3cm]{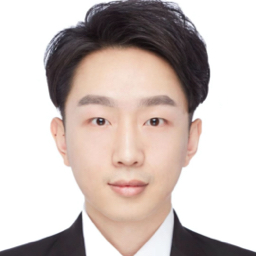}}},{}]
\noindent
\textbf{Lei Jin} is currently a Postdoc with the Beijing University of Posts and Telecommunications (BUPT), Beijing, China. He graduated from the same university with a
Ph.D. degree in 2020. He received the bachelor degree in the BUPT in 2015. His research
interests include network security, traffic security analysis, machine learning and pattern recognition, with a focus on 6Dof poses estimation and Human pose estimation.
\end{figwindownonum}
\vspace{1 cm}

\begin{figwindownonum}[0,l,{\mbox{\includegraphics[width=3cm]{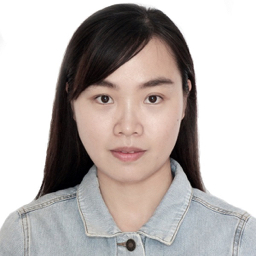}}},{}]
\noindent
\textbf{Nana Fan} is currently pursuing a Ph.D. degree in the Department of Computer Science, Harbin Institute of Technology, Shenzhen. She received a Master's degree from the Department of Computer Science, Harbin Institute of Technology, Shenzhen, in 2016. Her research interests include visual tracking, computer vision, and machine learning. 
\end{figwindownonum}
\vspace{1.2 cm}

\begin{figwindownonum}[0,l,{\mbox{\includegraphics[width=3cm]{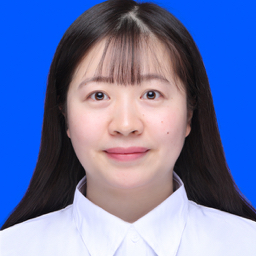}}},{}]
\noindent
\textbf{Min Wang} is currently pursuing the Ph.D. degree from Beijing Jiaotong University 
under the supervision of Prof. Congyan Lang. She received the B.E. degree from Beijing 
Information Science and Technology University, Beijing, China, in 2015. She has published over 4 cutting-edge papers on cross-modal/large-scale/semantic-based/image synthesis as the first author (including the following conferences and journals: IEEE MULTIMEDIA, ICME, TIST, IJCV). She was invited to give an “oral presentation” at ICME 2020. Her main research interests include multimedia information retrieval and analysis, machine learning, and computer vision.
\end{figwindownonum}
\vspace{0.1 cm}
 
\begin{figwindownonum}[0,l,{\mbox{\includegraphics[width=3cm]{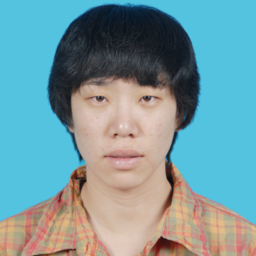}}},{}]
\noindent
\textbf{Xiaojuan Wang} received a Ph.D. degree in Electronic Science and Technology from 
the University of Beijing University of Posts and Telecommunications (BUPT). She is 
currently an associate professor in the same School of Electronic Engineering, Beijing, China. Her research interests include big data, complex networks, and sensors. Furthermore, she concentrates on using the six-axis sensor to do human activity recognition and the multi-source fusion in sensors. Her past Interests include wireless communication and software defined network.
\end{figwindownonum}
\vspace{0.5 cm}

\begin{figwindownonum}[0,l,{\mbox{\includegraphics[width=3cm]{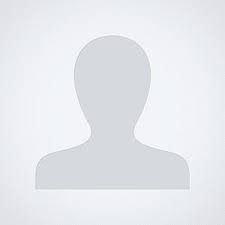}}},{}]
\noindent
\textbf{Ting Yong} is currently a Senior Engineer with National Key Laboratory of Science 
and Technology on Information System Security, Beijing, China. Her research interests 
include the security of intelligent unmanned system and wireless network protocol.
\end{figwindownonum}
\vspace{0.5 cm}

\begin{figwindownonum}[0,l,{\mbox{\includegraphics[width=3cm]{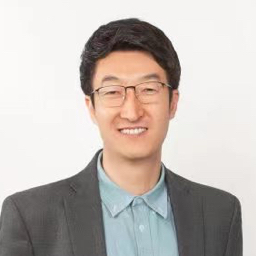}}},{}]
\noindent
\textbf{Yafeng Deng} is the vice president of 360's ARTIFICIAL intelligence Research Institute and head of its search business unit. He used to work in Baidu Deep Learning Institute, responsible for face recognition. He has 16 years of experience in computer vision and artificial intelligence. He has published more than 10 papers and obtained 95 national patents. He has 14 years of experience in artificial intelligence with a particular focus on computer vision. He has published more than 10 papers and obtained 95 patents. He used to be a scientist in Baidu Deep Learning Research Institute. He led the team to develop the world's first face detection and face recognition algorithm, and the developed algorithm and system serve hundreds of millions of user products.
\end{figwindownonum}

\begin{figwindownonum}[0,l,{\mbox{\includegraphics[width=3cm]{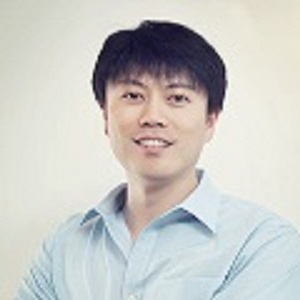}}},{}]
\noindent
\textbf{Yandong Guo}  is currently the OPPO Chief Scientist of Intelligent Perception and 
Adjunct Professor of Beijing University of Posts and Telecommunications. He gained his 
PhD from School of Electronic and Computer Engineering, Purdue University (West 
Lafayette) under the supervision of Prof. Jan P. Allebach and Charles A. Bouman. Dr. Guo 
mainly focuses on computer vision and artificial intelligence research and applies these 
research in industry applications. His papers have been widely accepted in CVPR, ECCV, 
ICML and other internationally recognized academic conferences and journals, and have 
been cited thousands of times by peers, and have also empowered many core products of 
companies including GE, HP, Microsoft, Xpeng Motors and OPPO. He has also served as a 
program committee member or reviewer for multiple conferences and journals and has 
organized ICCV and CVPR Workshops as the chair member.
\end{figwindownonum}

\begin{figwindownonum}[0,l,{\mbox{\includegraphics[width=3cm]{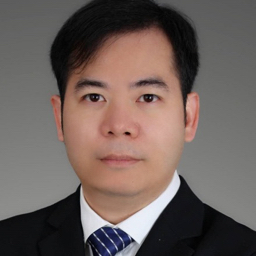}}},{}]
\noindent
\textbf{Shiming Ge} is a Professor with the Institute of Information Engineering, Chinese Academy of Sciences. Previously, he was a senior researcher and project manager in Shanda Innovations, a researcher in Samsung Electronics and Nokia Research Center. He received the B.S. and Ph.D degrees both in Electronic Engineering from the University of Science and Technology of China (USTC). His research mainly focuses on computer vision, data analysis, machine learning and AI security, especially efficient learning solutions towards scalable applications. He is a senior member of IEEE, CSIG and CCF. 
\end{figwindownonum}

\end{document}